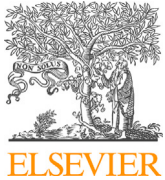
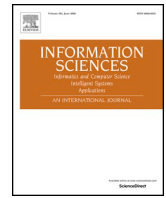

# Rolling the dice for better deep learning performance: A study of randomness techniques in deep neural networks

Mohammed Ghaith Altarabichi [*], Sławomir Nowaczyk, Sepideh Pashami, Peyman Sheikholharam Mashhadi, Julia Handl

*Center for Applied Intelligent Systems Research,[1] Halmstad University, Sweden*

## A B S T R A C T

This paper presents a comprehensive empirical investigation into the interactions between various randomization techniques in Deep Neural Networks (DNNs) and their impact on learning performance. It is well-established that injecting randomness into the training process of DNNs, through various approaches, at different stages, is often beneficial for reducing overfitting and improving generalization. Nonetheless, the interactions between randomness techniques such as weight noise, dropout, and many others remain poorly understood. Consequently, it is challenging to determine which methods can be effectively combined to optimize DNN performance. To address this issue, we categorize the existing randomness techniques into four key types: injection of noise/randomness at the data, model structure, optimization or learning stage. We use this classification to identify gaps in the current coverage of potential mechanisms for the introduction of randomness, leading to proposing two new techniques: adding noise to the loss function and random masking of the gradient updates.

In our empirical study, we employ a Particle Swarm Optimizer (PSO) for hyperparameter optimization (HPO) to explore the space of possible configurations to determine where and how much randomness should be injected to maximize DNN performance. We assess the impact of various types and levels of randomness for DNN architectures across standard computer vision benchmarks: MNIST, FASHION-MNIST, CIFAR10, and CIFAR100. Across more than 30 000 evaluated configurations, we perform a detailed examination of the interactions between randomness techniques and their combined impact on DNN performance. Our findings reveal that randomness through data augmentation and in weight initialization are the main contributors to performance improvement. Additionally, correlation analysis demonstrates that different optimizers, such as Adam and Gradient Descent with Momentum, prefer distinct types of randomization during the training process. A GitHub repository with the complete implementation and generated dataset is available.[2]

## 1. Background

Gradient Descent (GD) is a simple yet widely used algorithm to train Deep Neural Networks (DNNs). It computes the gradients of a loss function calculated based on the true labels and the network outputs during the feed-forward stage and uses them to update the DNN weights during backpropagation to minimize the loss function. The GD algorithm benefits from a number of techniques introducing randomness/noise at different stages of the process. This paper provides an overview of the practical usefulness of various choices and their combinations. To this end, we first categorize these randomization techniques into four distinctive classes: randomness in data, model, optimization, and learning.

*1.1. Randomness in data*

This class of approaches is agnostic towards the learning process, as randomness is introduced at the data level rather than within the algorithm itself. The general objective is to improve the diversity of the data, thereby reducing overfitting. Some of the most commonly used techniques in this category include:

– **Data shuffling**: Shuffling the data at each epoch is a common practice in training neural networks. Shuffling aims to break any order within the data and guarantees that batches will be different from one epoch to the next.
– **Data augmentation**: A common technique of applying random label-preserving transformations to the original data. For instance, in image data augmentation, images might be randomly rotated, flipped, or zoomed in during each training epoch or iteration [1]. This random variation diversifies the data seen by the model during training, preventing it from memorizing specific patterns and enhancing its ability to generalize to new, unseen data. By applying these random transformations at each training step, varying levels of randomness can be added to the training process.
– **Input noise**: The addition of small amounts of input noise has been found to improve the generalization performance in neural networks. Input noise can be regarded as a form of data augmentation, although typically applied at the pixel level, and has been shown to exert a smoothing effect on the structure of the input space [2]. Typically, normally distributed noise is added, and its level is controlled by the standard deviation.
– **Label smoothing**: Noise added to the labels is another regularization technique that can lead to better generalization [3]. This technique also improves model calibration by preventing the network from becoming overly confident about its own predictions.

*1.2. Randomness in models*

DNNs involve a number of structural components, and previous research has explored incorporating noise and randomness into the associated design decisions. Many of the following approaches are proposed to improve both training efficiency and act as regularization methods:

– **Weight initialization**: There are different strategies for setting the initial values of the weights and biases within a given network. Such initialization has been shown to have a significant effect on the convergence of DNNs training. A number of strategies exist to randomly initialize the weights and biases from a Gaussian or uniform distribution. The standard deviation of the initial distribution usually considers the number of inputs to the layer as in the He initialization method (commonly used with ReLU), and the Xavier initialization method (commonly used with Sigmoid and Tanh).
– **Dropout**: The key idea behind dropout is to randomly mask the output of a number of neurons during each feed-forward pass of training, making it fundamentally different from weight initialization, where randomization takes place once at the very beginning. Dropout has been used in many successful applications of deep learning due to its efficiency in reducing overfitting. However, the regularizing effect of dropout does not always protect against overfitting, as shown by Zhang et al. [4], who were able to perfectly fit a network regularized with dropout to the CIFAR10 dataset even when all the image labels were replaced with random ones.
– **DropConnect**: A generalization of dropout operates at the level of individual weights rather than entire neurons. On each feed-forward pass of the training, a random subset of weights is masked, meaning each unit receives input from a random subset of units in the preceding layer. This approach is commonly used to regularize fully connected layers [5].
– **Activation function noise**: The injection of noise into the output of the activation function has been shown to encourage GD to explore the seraph space of parameter values more. This trick has been found to be particularly effective at addressing the saturating behavior of some activation functions like Sigmoid and Tanh [6].
– **Loss function noise**: We propose an approach that has not yet been explored in the literature but may have a similar regularizing effect. *Loss function noise* introduces randomness to the loss function by multiplying it with a value sampled from a normal distribution, for every training instance. This approach allows the annealing of the noise magnitude as the training progresses.

*1.3. Randomness in optimization*

Stochastic Gradient Descent (SGD) is a variant of GD that computes an approximation of the gradient using a partial sample of training instances rather than using the entire dataset. Approximate optimization approaches of this kind have been shown to have a regularizing effect on the learning algorithm and usually lead to better generalization on the test set [7]. The approximate calculations of the gradient introduce random fluctuations in the training dynamics of SGD [8], the magnitude of which is known as the "noise scale" $g$, which is defined as:

$$g = \epsilon_{\text{eff}}(\frac{N}{B} - 1), \tag{1}$$

where $\epsilon_{\text{eff}}$ is the effective learning rate, $N$ is the number of training instances and $B$ is the batch size. The effective learning rate in the case of SGD and SGD with Momentum can be expressed as $\epsilon_{\text{eff}} = \frac{\epsilon}{1-m}$, where $\epsilon$ is the initial learning rate, and $m$ is the momentum. $g$ can then be approximated for a small batch size $B \ll N$ as:

$$g \approx \frac{\epsilon N}{B(1-m)}. \tag{2}$$

It is clear from Equations (1) that $g \to 0$ when batch training ($B = N$) is used, which intuitively indicates noise-free optimization as the entire dataset is used to calculate the gradient. A number of techniques and hyperparameters interact, affect, and control the noise scale:

- **Optimizer**: The optimization algorithm is the core component that drives learning in DNNs. The main difference between different families of optimizers is the way the effective learning rate $\epsilon_{\text{eff}}$ is computed. The effective learning rate $\epsilon_{\text{eff}}$ for vanilla SGD ($m = 0$) is $\epsilon_{\text{eff}} = \epsilon$, therefore, the noise scale is $g \approx \frac{\epsilon N}{B}$. On the other hand, for Gradient Descent with Momentum, a larger value of the momentum ($m > 0$) will increase the value of the effective learning rate $\epsilon_{eff}$ and consequently increase the noise scale $g$ based on Equation (2). The family of adaptive optimizers, e.g., Adam is more complex as the adaptive learning rate is computed for each parameter not only based on $m$, but also based on the exponentially decaying average of past squared gradients for the parameters.
- **Learning rate**: The learning rate is an important hyperparameter that affects the speed at which the network learns. The noise scale $g$ of SGD explicitly depends on the learning rate according to Equation (2). It must be noted that the effect of the learning rate is not limited to varying the noise, as a large learning rate would not only increase the noise but would also encourage overshooting [9].
- **Learning rate scheduler**: The common practice of decaying the learning rate according to a certain factor after a pre-defined number of epochs is equivalent to annealing the noise in training neural networks, as evident from Equation (2). It is widely accepted that a large initial learning rate with a decaying schedule improves performance comparing to the use of a constant learning rate [10].
- **Mini-batch**: The partial sample of training instances used by SGD to approximate the gradient is known as the mini-batch. The batch size is usually chosen between one and a few hundred samples.
- **Batch size scheduler**: Increasing the batch size over the course of training has been found to improve test accuracy with fewer parameter updates per epoch [8]. From a noise perspective, the technique has a comparable effect to a learning rate decay (see Equation (2)) as it gradually reduces the noise scale in the gradient estimates.
- **Weight noise**: Gaussian noise added to the weights in every training step is shown to have a simplifying and regularizing effect on the network [11]. However, some training instability has been observed when weight noise is added [12].
- **Gradient noise**: Annealed Gaussian noise is added to the gradient in every training step. This technique is different from adding noise directly to the weights in the case of adaptive or momentum-based optimizers, where $\epsilon_{\text{eff}}$ also relies on prior training steps. This approach is found to be particularly useful for deep models and in special cases of poor weight initialization [12].
- **Gradient Dropout**: We propose to extend the *Gradient Dropout* idea originally proposed in the context of meta-learning [13] to the supervised training setting. The concept is based on regularizing the network by randomly masking some of the gradient updates during backpropagation.

*1.4. Randomness in learning*

Randomized learning techniques offer an alternative paradigm to the conventional backpropagation-trained neural networks. This is often based on the "Randomization is cheaper than optimization" philosophy. In particular, these techniques propose to randomly initialize, and then fix, most network weights and biases. The main goal is to streamline the optimization process and enable quicker training by solving the optimization problem exactly (or almost exactly), compared to the iterative parameter updates of backpropagation. We focus here on learners who embrace randomization as an intrinsic feature of the learning process:

- **Randomized Neural Networks**: The fundamental concept behind this type of learning is based on keeping the majority of weights, typically all but the final layer, fixed throughout the training. The training process is simplified, with optimization efforts concentrated on solving output weights using closed-form solutions (e.g., standard linear least squares). The ideas of fixing weights date back to the original perceptron architecture. The authors in [14] show that for a Single-Layer Feed-Forward Neural Network (SLFN), numerous parameter solutions often yield good performance, with the selection made by the output layer being more significant than the weights of hidden units. Accordingly, assigning random weights to hidden units and using the Fisher method to determine the weights for the output units is a viable approach. The underpinning phenomenon of stochastic separability of random points in high dimension was later confirmed by [15], showing that linear discrimination can be explicitly found using Fisher method with a probability close to one.
A prominent example of this approach are the Random-Vector Function Link (RVFL) nets introduced by Pao et al. [16], operating as SLFN. RVFL networks are proven to be universal approximators of continuous functions [17]. However, the practical success of RVFL networks is dependent on the "proper" selection of random weights [18]. Several randomized learner models such as Stochastic Configuration Networks (SCN) [19] and Stochastic Configuration Machines (SCM) [20] offer supervisory mechanisms for assigning random weights and selecting their scope to ensure the network's universal approximation property. There are recent attempts to extend these SLFNs to deeper architectures like Deep Stochastic Configuration Networks (DeepSCNs) [21] and deep RVFL (dRVFL) [22].
Similarly, in the domain of time series data modeling, Echo State Networks (ESNs) [23] have emerged. They are recurrent randomized neural networks leveraging the concept of reservoir computing, i.e., high-dimensional mapping utilizing the dynamics of a fixed, non-linear system. ESNs feature untrained recurrent layers and have been shown to yield excellent performance in many benchmarks tasks without full adaptation of all network weights [24,25]. Recent advancements include the exploration of stacked recurrent layers to create more expressive and deeper architectures, as demonstrated by approaches like deepESN [26], and DeePr-ESN [27].

**Table 1**
Popular randomization techniques in DNN categorized by purpose.

| Category | Technique | Regularization | Data size | Convergence | Training Time |
|---|---|---|---|---|---|
| Data | Data shuffling | | ✓ | | |
| | Augmentation | ✓ | ✓ | | |
| | Input noise | ✓ | ✓ | | |
| | Label smoothing | ✓ | | | |
| Model | Weight initialization | | | ✓ | |
| | Dropout | ✓ | | | |
| | DropConnect | ✓ | | | |
| | Activation noise | | | ✓ | |
| Optimization | Learning rate | ✓ | | ✓ | ✓ |
| | Learning rate schedule | ✓ | | ✓ | ✓ |
| | Mini-batch | ✓ | | ✓ | ✓ |
| | Batch schedule | | | ✓ | ✓ |
| | Weight noise | ✓ | | ✓ | |
| | Gradient noise | | | ✓ | |
| Learning | Randomized Neural Networks | | | ✓ | ✓ |

*1.5. Summary on randomization techniques and main contributions*

As highlighted in Table 1, all these techniques have been designed in pursuit of common goals: regularization, dealing with limited data, improving convergence, and training efficiency. Nevertheless, they are generally studied independently, with very limited work on potential interactions between them. We conduct an ablation study addressing this gap by using a meta-heuristic optimizer, Particle Swarm Optimization (PSO), to perform a hyperparameter search to identify "good" randomness settings in computer vision tasks.

We compare the individual contributions and their interactions with regard to the overall performance gain. Our objective is to identify compatible well-performing methods and to recognize mutually redundant mechanisms (e.g., an increasing batch size is similar to a learning rate decay) and hard-to-optimize ones (e.g., weight noise is associated with training instability). Our contributions are as follows:

- We deploy a PSO hyperparameter algorithm to comprehensively optimize the levels of randomness for 22 techniques in DNNs. Our approach achieved an average a 38.4% reduction in test error rates across various vision benchmark tasks.
- Our analysis of the top-performing settings shows that different optimizers, such as Adam and Gradient Descent with Momentum, work best with distinct mechanisms for the randomization of the training process.
- An ablation study reveals that data augmentation and weight initialization are the top contributors to performance improvement in CNNs.
- By categorizing randomness techniques into four types impacting at the data, model, optimization, and learning levels, we identify gaps in current coverage and propose two new techniques: loss noise and gradient dropout.

The remainder of the article is organized as follows. In Section 2, we discuss related work on studying and tuning randomness techniques in DNNs. In Section 3, we explain our hyperparameter optimization approach and define the two proposed new randomization techniques: loss function noise and gradient dropout. The experiments are presented in Section 4, and the results are further discussed in Section 5. Conclusions are listed in Section 6, while limitations and future work are covered in Section 7.

**2. Related work**

Our related work section is organized into two parts. Firstly, we discuss the literature analyzing and studying the role of randomness in training DNNs. In the second subsection, we discuss popular approaches to hyperparameter optimizations in DNNs.

*2.1. The role of randomness in deep neural networks*

The early work of studying the role of randomness in neural network training dates back to [28], where the authors analyzed the effect of introducing additive noise to the inputs. Similar results about the usefulness of adding noise to the input for classification and regression tasks were reported in [2]. The same study found weight noise to be successful only in classification tasks, with no improvement in generalization observed for output noise in their settings. The effectiveness of weight noise is further questioned in [12], where models failed to train even with weight noise rates as low as $10^{-6}$.

The use of noise in training DNN is known to cause a number of tradeoffs. Probably the best well-known is the underfitting/overfitting tradeoff, as observed in stochastic regularization methods like Dropout. An increased level of randomness (e.g., Dropout rate) can lead to poor data fitting, while too little noise weakens regularization and can cause overfitting. Another tradeoff exists between computation time and generalization. Using a large batch size reduces the noise in the gradient estimate and can significantly reduce training time. However, this reduction in noise may lead to poor generalization due to an increased chance of converging to a sharp minimum [7]. The problem of small datasets can be eased with image augmentation, although there is a risk of unsafe transformations that do not preserve the label (e.g., random erasing can turn digit 8 to 6). Additionally, the injection of noise into the optimization during training can encourage the optimizer to escape local optima [9] but this approach is often associated with training instabilities [12].

Albeit with a different objective from our work, the role of randomness in training DNN is also studied in [29], particularly from the perspective of replicability. A comprehensive experimental study is carried out to characterize the non-determinism in DNN training using a number of architectures, datasets, and software libraries in computer vision. The authors found that the presence of stochasticity due to algorithmic and implementation choices can lead to significant variations in SotA model performance.

*2.2. Hyperparameter optimization*

The optimization of randomness in neural networks can be thought of as a Hyperparameter Optimization (HPO) problem. This problem is in equal parts art and science, with the common consensus in the deep learning community that different datasets, tasks, and families of learning algorithms can require different settings. An important aspect is the construction of a search space that defines the hyperparameters to search and the limits of the search [30]. A popular HPO choice among practitioners is Grid Search (GS), which has been successfully applied to a number of Machine Learning (ML) HPO problems [31] with a small number of hyperparameters, often assumed to be $\leq 5$. As an exhaustive search method, GS suffers from the curse of dimensionality, as the complexity of brute-forcing combinations of values grows exponentially with more hyperparameters to tune. The high dimensionality problem is particularly evident for DNNs which involve a large number of hyperparameters and algorithm design choice.

A straightforward alternative to GS is Random Search (RS), which is widely established as the baseline search procedure for HPO problems for DNNs. RS usually outperforms GS for large search spaces and finds solutions faster. However, both methods share the limitation of sampling independently of previous evaluations. Therefore, these methods are not capable of fully exploiting well-performing regions of the search space [32].

A wide variety of optimization approaches are used with HPO problems of complex search spaces, including PSO [33], Genetic Algorithms (GA) [34], Bayesian methods [35] and the Iterated Racing Procedure [36]. Population-based approaches like GAs are generally found to outperform GS and RS, particularly with large search spaces [37]. Similarly, PSO has been shown to effectively explore the solution space of DNN hyperparameters, achieving a DNN with competitive performance despite a minimal architecture as shown in [38]. PSO has also been shown to improve the performance of CNN's with the potential to benefit from parallelization [39] to reduce the search time further.

**3. Method**

The common goal of a Machine Learning (ML) task $\mathcal{T}$ is to construct a model $\mathcal{M}_\theta$ using a learning algorithm $\mathcal{A}$ that minimizes a predefined loss function $\mathcal{L}(X^{te}; \mathcal{M}_\theta)$ over a set of unseen test data $X^{te}$. This goal is achieved by training $\mathcal{M}_\theta$ via optimizing its parameters $\theta$ to minimize $\mathcal{L}(X^{tr}; \mathcal{M}_\theta)$ for a given training set $X^{tr}$ of labeled input-output pairs. A validation dataset $X^{va}$ is often set aside to estimate $\mathcal{L}(X^{te}; \mathcal{M}_\theta)$ by monitoring $\mathcal{L}(X^{va}; \mathcal{M}_\theta)$ during training. However, $\mathcal{A}$ often has a number of hyperparameters $\lambda$ whose values must be chosen prior to training $\mathcal{A}(X^{tr}; \lambda)$. The objective of a hyperparameter search is to identify the set of hyperparameters $\lambda^*$ that can lead to an optimal model $\mathcal{M}_\theta = \mathcal{M}_\theta^*$ minimizing $\mathcal{L}(X^{va}; \mathcal{M}_\theta)$. This can be formalized as:

$$\lambda^* = \arg\min_\lambda \mathcal{L}(X^{va}; \mathcal{A}(X^{tr}; \lambda)) = \arg\min_\lambda \mathcal{F}(\lambda; \mathcal{A}, X^{tr}, X^{va}, \mathcal{L}). \tag{3}$$

The objective function $\mathcal{F}$ accepts a candidate hyperparameter setting $\lambda$ and returns the corresponding $\mathcal{L}$ on the validation split $X^{va}$. The definition of $\mathcal{F}$ in Equation (3) can of course be directly extended to the maximization of a performance measure (metric) such as accuracy $\mathcal{U}$:

$$\lambda^* = \arg\max_\lambda \mathcal{F}(\lambda; \mathcal{A}, X^{tr}, X^{va}, \mathcal{U}). \tag{4}$$

A metaheuristic is often used to carry out the optimization guided by $\mathcal{F}$ in Equation (4). In our work, we use PSO, a popular stochastic optimization method that randomly initializes a swarm of particles and searches the space through their movements guided by their awareness of the current, globally best position of any member of the swarm. In the context of HPO, the position of a particle $i$ at iteration $t$ in the d-dimensional space $\mathbb{R}^d$ represents the hyperparameter setting $\lambda_{i,t} \in \mathbb{R}^d$, where $d$ represents the number of hyperparameters of the HPO problem. At each iteration, the updates of the velocity $V_{i,t+1}$ are calculated according to:

$$V_{i,t+1} = wV_{i,t} + c_1\varphi_{1i,t}(p_{i,t} - \lambda_{i,t}) + c_2\varphi_{2i,t}(g_t - \lambda_{i,t}), \tag{5}$$

where $w$ is the inertia weight affecting the contribution of the prior velocity, $c_1$ and $c_2$ are the social and global acceleration coefficients, respectively, controlling how much the velocity is affected by the local $p_{i,t}$ and the global $g_t$ best positions. $\varphi_{1i,t}$, $\varphi_{2i,t}$ are random variables sampled in each iteration $t$ for each particle $i$ from a uniform distribution between 0 and 1 to introduce randomness to the contributions of the acceleration coefficients.

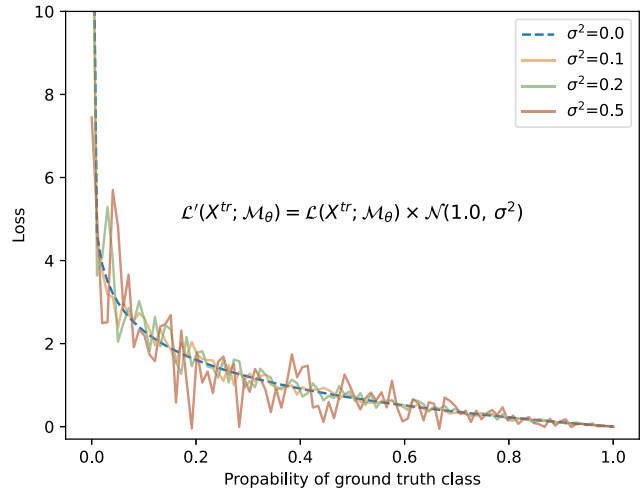

**Fig. 1.** The noisy version of the log loss for varying values of loss noise.

$$\lambda_{i,t+1} = \lambda_{i,t} + V_{i,t+1} \tag{6}$$

This optimization is carried out iteratively until the stopping criterion (defined either by a total number of iterations, evaluations, or the number of iterations with no progress in terms of the best fitness) is met. Eventually, the optimization returns $\lambda^*$, the set of hyperparameters with the best $\mathcal{F}$ value found.

The search for the optimized setting $\lambda^*$ for task $\mathcal{T}$ is the first step in our ablation procedure. Once $\lambda^*$ is identified, we proceed by removing its decision variables one at a time (by setting its value to the default e.g., dropout rate is zero) to measure their individual contributions to the fitness $\mathcal{F}(\lambda^*; \mathcal{A}, X^{tr}, X^{te}, \mathcal{U})$. We measure $\mathcal{U}^k$, the fitness contribution of the decision variable $k$, by creating the setting $\lambda^k$ defined as:

$$\lambda^k = \begin{cases} \lambda^k[j] = \lambda^0[j], & \text{if } j = k \\ \lambda^k[j] = \lambda^*[j], & \text{otherwise} \end{cases}, \tag{7}$$

where $\lambda^0$ is the default setting, $j \in \{1, 2, ..., d\}$ is the decision variable index. Then we calculate $\mathcal{U}^k$ as:

$$\mathcal{U}^k = \mathcal{F}(\lambda^*; \mathcal{A}, X^{tr}, X^{te}, \mathcal{U}) - \mathcal{F}(\lambda^k; \mathcal{A}, X^{tr}, X^{te}, \mathcal{U}) \tag{8}$$

Assuming a maximization problem, a positive value $\mathcal{U}^k$ value indicates that performance drops when we remove the decision variable $k$ from $\lambda^*$ according to Equation (8). We calculate $\mathcal{U}^k$ for $k \in \{1, 2, ..., d\}$ to quantify the individual contributions of each decision variable in our study. Based on this, we can calculate the Error Reduction Rate (ERR), to quantify improvements in model performance for the decision variable $k$ as:

$$ERR^k = \frac{\mathcal{U}^k}{\mathcal{U}^{max} - \mathcal{F}(\lambda^k; \mathcal{A}, X^{tr}, X^{te}, \mathcal{U})}, \tag{9}$$

where $\mathcal{U}^{max}$ is the maximum possible value of the metric $\mathcal{U}$ (for example, 100% for accuracy).

### 3.1. Loss noise and gradient dropout

In this section, we formalize the definitions for our two novel proposals of randomness techniques: loss function noise and gradient dropout.

The noisy value of the loss function, $\mathcal{L}'(X^{tr}; \mathcal{M}_\theta)$, is calculated based on the predefined loss function $\mathcal{L}(X^{tr}; \mathcal{M}_\theta)$ according to the following formula:

$$\mathcal{L}'(X^{tr}; \mathcal{M}_\theta) = \mathcal{L}(X^{tr}; \mathcal{M}_\theta) \times \mathcal{N}(1.0, \sigma^2) \tag{10}$$

where $\mathcal{N}$ is a normally distributed noise with a mean of 1, and $\sigma$ is the standard deviation that controls the magnitude of noise introduced to the loss calculations. We illustrate the effect of this noisy variant of the Cross-Entropy (CE) loss in Fig. 1.

Gradient dropout provides a mechanism for introducing stochasticity to gradient updates, similar to how dropout introduces stochasticity to activations during forward passes. It involves randomly setting the gradients of some parameters to zero during each optimization step.

We denote the gradient of the loss function with respect to the parameters $\theta$ as $\nabla_\theta L(\theta)$. During optimization, for each parameter $\theta_i$, a binary dropout mask $d_i$ is sampled independently from a Bernoulli distribution with a probability $p$ of being 1 (indicating that the gradient is retained) and $1 - p$ of being 0 (indicating that the gradient is dropped). The masked gradient update $\Delta \theta_i$ is then computed for each parameter as:

$$\Delta \theta_i = d_i \cdot \nabla_{\theta_i} L(\theta) \tag{11}$$

During training, gradient dropout masks are sampled for each mini-batch and are applied independently to each parameter.

## 4. Experiments

The insights presented in this paper arise from two experiments; in the initial one, we select a reliable search method and use it to identify successful randomness settings. Accurately determining high-performing configurations is fundamental, since, further on, we draw conclusions from the frequency, individual impacts, and co-occurrences of the techniques within the selected settings. In the second experiment, we follow the ablation procedure to identify the most successful randomness techniques based on their performance in reducing errors, across various types of networks and benchmark datasets. We opted to carry out an ablation study as it is a popular procedure of uncovering complex interaction within DNNs. Many examples from the computer vision literature showcase its effectiveness across various domains including in: image classification [40], object and face detection [41], HPO and architectural components [42].

We use four popular datasets for our experiments. For each of them, we set aside 20% of the training data as a validation split to evaluate the fitness during the search. **MNIST** is a handwritten digits classification dataset containing 60 000 28x28 black and weight images of 10 single digits, with 10 000 images reserved for testing. **Fashion-MNIST** is a fashion products dataset from Zalando Research, proposed as a direct complement for MNIST with the same image size, number of training and test samples, and a number of classes. **CIFAR-10** is a dataset created by the Canadian Institute for Advanced Research containing 60 000 32x32 color images in 10 different classes with the same training and test splits. **CIFAR-100** dataset is analogous to CIFAR10, except it contains 100 classes.

We evaluate two network architectures. **FC** is a simple fully connected network with two hidden layers of fifty neurons each. Each layer is followed by Dropout, DropConnect, and a Noisy ReLU. **CNN** is a VGG-like architecture with three blocks, each block consisting of two Conv2D followed by MaxPooling2D, Dropout, DropConnect, Noisy ReLU, and Batch Normalization layers, with 64, 128, and 256 channels, respectively. The classifier is one dense layer with 128 neurons, followed by Dropout and DropConnect. The overall number of parameters is 435 306.

We establish **Baseline** as the performance of the network trained using the default values listed in Table 2, for 22 parameters representing all of the different randomness mechanisms covered in Section 1. For the **PSO**, we use the following parameters: {$c_1$: 1.49618, $c_2$: 1.49618, $w$: 0.7298}, and a velocity clamp of 0.2 of the search space range. The number of particles is set to 44 (corresponding to 200% of the decision variables, as we followed the procedure described in [43]) in all experiments. The search terminates when no progress in the best solution is observed for 5 steps.

### 4.1. Experiment I - search procedure

We have considered three procedures for performing the HPO search: random search, separate optimization of each individual hyperparameter, and PSO. We immediately exclude a **Full Grid Search** because it is intractable; to optimize all 22 randomness techniques at once requires $10^{20}$ evaluations, each taking 10 minutes to train, for a total time in the order of 100 000 multiples of the age of the universe.

Therefore, we first use **Random Search** (RS) to uniformly and independently sample settings for all the randomness techniques within the ranges from Table 2. The search returns the best-found configuration after exhausting a budget of 1 000 total fitness evaluations. Next, we apply grid search separately to each parameter, ignoring interactions, and refer to this procedure as **Single-Technique Search** (STS). The discrete parameter ranges are exhaustively evaluated, while continuous ones are divided into ten equidistant values, resulting in a total of 200 evaluations.

The complete results are presented in Table 3. The results for RS clearly indicate the complexity of the search space. The network failed to train for the majority (63.9%) of the evaluated settings, and even the best solution found did not surpass the baseline performance. Clearly, RS is unable to identify promising regions of the search space and to exploit this knowledge efficiently by exploring them more in-depth, due to each evaluated sample being generated independently. This observation aligns with the "Edge of Chaos" characterizations in the literature, which highlight that, for deep networks, only very specific choices of hyperparameters lead to good performance [44].

The STS results in the same table show that the most successful single hyperparameter (input noise) optimization achieved 27.07% ERR over the default setting. At the same time, in the five runs, PSO achieved ERRs that ranged from 38.72% to 45.11%. All settings found by PSO recorded a higher ERR median (calculated over five independent runs of the same configuration) than any other setting found by RS and STS, which was statistically significant at 0.05 based on an independent Student's *t*-test. This result supports our intuition that the right combination of randomness techniques improves performance more than what could be achieved with one of the underpinning individual techniques alone. Thus, we will use PSO for HPO throughout the rest of the paper, specifically to carry out the ablation study.

**Table 2**
Randomization techniques included in our ablation study with types, default values, and acceptable ranges.

| Technique | Default Value | Parameter Range |
| --- | --- | --- |
| Data shuffling | False | [False, True] |
| Random flip | none | [none, horizontal, vertical, both] |
| Random rotation | 0.0 | [0.0, 1.0] |
| Random zoom | 0.0 | [0.0, 1.0] |
| Random translation | 0.0 | [0.0, 1.0] |
| Random contrast | 0.0 | [0.0, 1.0] |
| Input noise | 0.0 | [0.0, 1.0] |
| Label smoothing | 0.0 | [0.0, 1.0] |
| Weight initialization | 0.05 | [0.0, 0.1] |
| Dropout | 0.0 | [0.0, 0.5] |
| DropConnect | 0.0 | [0.0, 0.5] |
| Deep Randomized Neural Networks[*] | 0.0 | [0.0, 1.0] |
| Activation noise | 0.0 | [0.0, 0.1] |
| Loss noise | 0.0 | [0.0, 1.0] |
| Optimizer | Vanilla SGD | [Vanilla SGD, SGD with momentum, Adam] |
| Learning rate | 0.001 | [0.1, 0.00001] |
| Learning rate scheduler | 0.0 | [-1.0, 1.0] |
| Mini-batch | 128 | [128, 1024] |
| Batch size scheduler | 0.0 | [-1.0, 1.0] |
| Weight noise | 0.0 | [0.0, 0.000001] |
| Gradient noise | 0.0 | [0.0, 1.0] |
| Gradient dropout | 0.0 | [0.0, 0.5] |

[*] DRNN in our experiments is an adaptation of the Randomized Neural Network concept of fixing weights throughout the training. We introduce a configurable parameter that controls the percentage of the fixed weights within the hidden layers. To remain consistent with the overall experimentation framework, we have used error backpropagation to learn the unfixed weights.

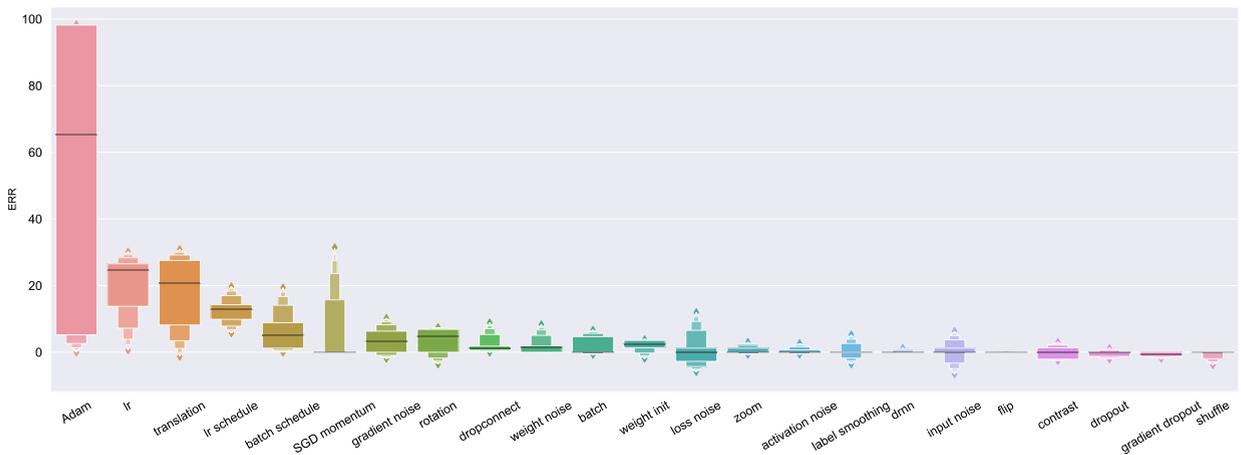

**Fig. 2.** The randomness techniques ranked based on average ERR contributions for the five optimized runs of the FC network and the MNIST dataset.

*4.2. Experiment II - the ablation study*

In this experiment, we apply our ablation procedure to the five optimized settings of the FC network using the MNIST dataset from Table 3. The boxplots in Fig. 2 show the distributions of the ERR results of all 22 techniques ranked by average ERR. In total, four techniques recorded a positive ERR median over the five runs, statistically significant at 0.05 based on the independent Student's *t*-test. The most significant factor in improving performance was the choice of the optimizer. Adam was chosen four times out of the five runs, and replacing it with Vanilla SGD in the ablation procedure resulted in training instabilities, particularly with large batch sizes. The other top median ERR contributions came from learning rate, random translation, and learning rate scheduler with 32.52%, 26.0%, and 14.72%, respectively. A decaying learning rate scheduler and an increasing batch scheduler were always chosen by the optimizer. Surprisingly, the Dropout technique had poor performance, offering a positive ERR contribution only once. PSO seemed to consistently favor the use of the other variant, DropConnect, which offered a positive ERR contribution in four of the five settings.

We tested more scenarios for the CNN network, compared to FC, by including all four datasets: MNIST, Fashion MNIST, CIFAR10, and CIFAR100. As observed from Table 4, all 20 HPO runs returned randomness settings that significantly improved upon the baseline performance according to the independent Student's *t*-test. The ERR contributions ranged from 15.09% up to 48.07%. The ablation results are aggregated in Fig. 3, showing the ERR contributions of all 22 randomness techniques to all the 20 PSO settings from Table 4, ranked by the average. The highest ERR improvement comes from the learning rate, widely considered as the single most important hyperparameter of DNNs [45]. Overall, the methods that achieved a positive ERR median over the 20 settings, statistically significant at 0.05 according to independent Student's *t*-test, are random flip, random translation, label smoothing, weight initialization, dropout, learning rate, and learning rate scheduler. An interesting setting was discovered during the search in MNIST with a weight initialization set to zero by the PSO optimizer. The weight symmetry in this setting is broken with activation noise, allowing training while using a zero constant initialization of the weights.

**Table 3**
Results of the FC network optimizations for MNIST dataset with STS, RS, and PSO. Results are reported as median of five independent runs, bold indicates statistical significance compared to default settings and italic over STS.

| Method | Test Accuracy | Error Rate Reduction |
|---|---|---|
| **Baseline** | | |
| Default settings | 97.34% | n/a |
| **Single-Technique Search** | | |
| Data shuffling | 97.46% | 4.51% |
| Random flip | **95.98%** | **-51.13%** |
| Random rotation | **97.57%** | **8.65%** |
| Random zoom | **97.85%** | **19.17%** |
| Random translation | **97.58%** | **9.02%** |
| Random contrast | 97.31% | -1.13% |
| Input noise | **98.06%** | **27.07%** |
| Label smoothing | **97.63%** | **10.9%** |
| Weight initialization | 97.22% | -4.51% |
| Dropout | **97.66%** | **12.03%** |
| DropConnect | **97.79%** | **16.92%** |
| DRNN | 97.37% | 1.13% |
| Activation noise | 97.54% | 7.52% |
| Loss noise | 97.43% | 3.38% |
| Optimizer | 97.16% | -6.77% |
| Learning rate | 97.6% | 9.77% |
| Learning rate scheduler | 97.42% | 3.01% |
| Mini-batch | 97.18% | -6.02% |
| Batch size scheduler | 97.29% | -1.88% |
| Weight noise | 97.35% | 0.38% |
| Gradient noise | 97.35% | 0.38% |
| Gradient dropout | 97.26% | -3.01% |
| **Random Search** | | |
| Optimized settings | 93.28% | -152.63% |
| **PSO** | | |
| Optimized settings I | ***98.54%*** | ***45.11%*** |
| Optimized settings II | ***98.5%*** | ***43.61%*** |
| Optimized settings III | ***98.45%*** | ***41.73%*** |
| Optimized settings IV | ***98.37%*** | ***38.72%*** |
| Optimized settings V | ***98.37%*** | ***38.72%*** |

## 5. Discussion

We summarize the above results as several conclusions that provide insights into the effects of randomness in the training and performance of neural networks.

### 5.1. Which category of randomness is the most important for CNN?

We may observe from Fig. 4 that randomness techniques under the data category contributed the most to performance improvements. This finding supports the unanimous agreement that performance improvements realized by DNNs can be mostly attributed to larger datasets [46]. This observation, however, should not be viewed as one denying the importance of the model and optimizer decisions. The top-performing configurations incorporated various mechanisms beyond data perturbations, emphasizing the additional benefits derived from noise injections at different levels of the training process.

**Table 4**
Results of the CNN network optimizations for all datasets. Results are reported as median of five independent runs.

| Dataset | Settings | Test Accuracy | Error Rate Reduction |
| --- | --- | --- | --- |
| MNIST | Baseline | 99.42% | n/a |
|  | PSO I | 99.69% | 46.55% |
|  | PSO II | 99.66% | 41.38% |
|  | PSO III | 99.66% | 41.38% |
|  | PSO IV | 99.63% | 36.21% |
|  | PSO V | 99.6% | 31.03% |
| Fashion MNIST | Baseline | 93.04% | n/a |
|  | PSO I | 94.58% | 22.13% |
|  | PSO II | 94.32% | 18.39% |
|  | PSO III | 94.25% | 17.39% |
|  | PSO IV | 94.14% | 15.81% |
|  | PSO V | 94.09% | 15.09% |
| CIFAR10 | Baseline | 79.26% | n/a |
|  | PSO I | 89.23% | 48.07% |
|  | PSO II | 88.84% | 46.19% |
|  | PSO II | 88.71% | 45.56% |
|  | PSO IV | 88.67% | 45.37% |
|  | PSO V | 88.64% | 45.23% |
| CIFAR100 | Baseline | 45.4% | n/a |
|  | PSO I | 62.01% | 30.42% |
|  | PSO II | 61.43% | 29.36% |
|  | PSO II | 61.24% | 29.01% |
|  | PSO IV | 60.36% | 27.4% |
|  | PSO V | 60.27% | 27.23% |

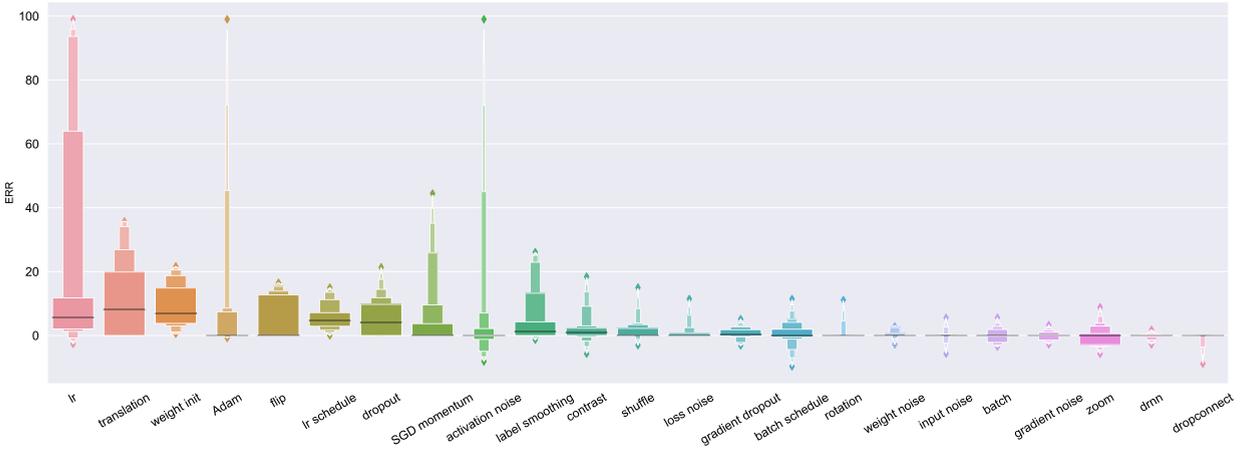

**Fig. 3.** The randomness techniques ranked based on ERR contributions for the 20 optimized runs of the CNN network and the four datasets MNIST, Fashion MNIST, CIFAR10, CIFAR100.

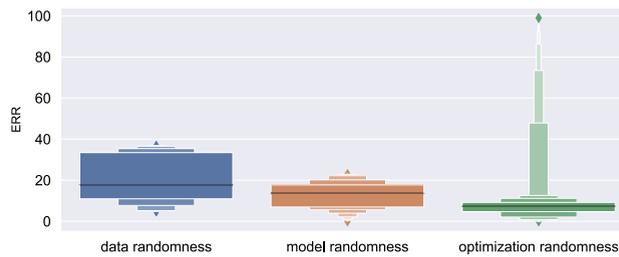

**Fig. 4.** Ablation results summarized at category level.

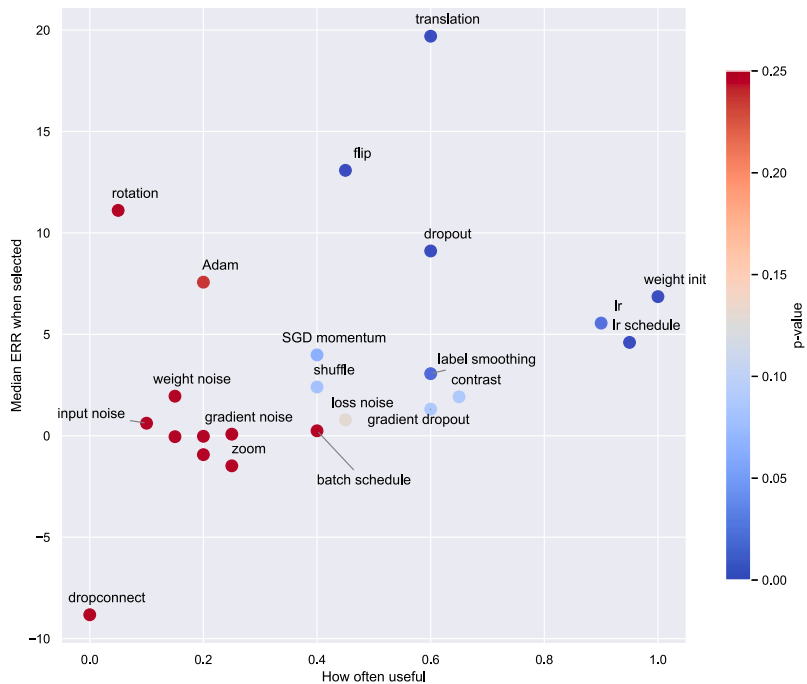

**Fig. 5.** Visualization of randomness techniques, by comparing the number of times each one achieved a positive ERR to its median ERR, i.e., how often versus how much each technique contributed (across 20 ablation runs, for the CNN network, all four datasets). Colors indicate the p-value of Student's *t*-test for the null hypothesis of "technique has median ERR $\leq 0$."

### 5.2. Which randomness technique is the most important for CNN?

The ranking previously presented in Fig. 3 may not provide a clear indication of how consistently a particular method performed across the 20 runs. To address this, we suggest to measure the technique's consistency by counting how often it exhibited positive ERR contributions. This approach enables us to identify techniques that consistently contributed to performance improvements across multiple scenarios (datasets). We propose, accordingly, to evaluate the importance of the techniques' contributions based on two objectives, the consistency of the technique and the median ERR of all contributions during the ablation procedure.

The multi-objective view presented in Fig. 5 suggests that weight initialization showed the most consistent performance (shown normalized, i.e., divided by 20), contributing positively in all 20 settings, followed by learning rate decay in 19, learning rate in 18, and random contrast in 13 settings, respectively. Dropout, label smoothing, and gradient dropout all contributed in 12 settings. Interestingly, the non-dominated front has two randomness techniques only, random translation as the most impactful technique based on median ERR, and weight initialization as the most consistently useful randomization approach.

### 5.3. Which randomness techniques are dataset-specific?

We analyze the interactions between randomness techniques by examining associations within the top-performing settings that resulted from our CNN ablation experiments. Throughout the HPO, we tested over 30,000 unique noise settings. For our analysis, we have selected the top 120 settings (best 0.39%) based on accuracy from each dataset/optimizer combination; the resulting heatmap, summarizing all these results, is presented in Fig. 6.

As expected, all data augmentation techniques showed clear dependency on the dataset. This association is most intuitively understandable for the random flip in MNIST data, given that a horizontal/vertical flip of a digit is, in many cases, not a label-preserving (safe) transformation. Unsurprisingly, the random flip was selected exclusively for CIFAR10 and CIFAR100. Random zoom, on the other hand, was only useful with MNIST and Fashion MNIST. Random contrast showed the most consistent performance, contributing positively in 13 settings, and was useful at least once for each dataset. Random translation, identified as a top contributor based on our ablation results, was used significantly more with CIFAR10 and CIFAR100, as shown in Fig. 7. These observations highlight the data-specific challenge of designing a generalizable augmentation policy [1].

The familiar regularization techniques like label smoothing and dropout showed significant dependence on the complexity of the dataset. This association could be attributed to the necessity for stronger regularization on simpler datasets, as they are more prone to overfitting. Interestingly, the same significant association was observed for the learning rate in Fig. 7. The regularizing effect of an initial large learning rate is known [10], as it prevents the memorization of noisy data [47].

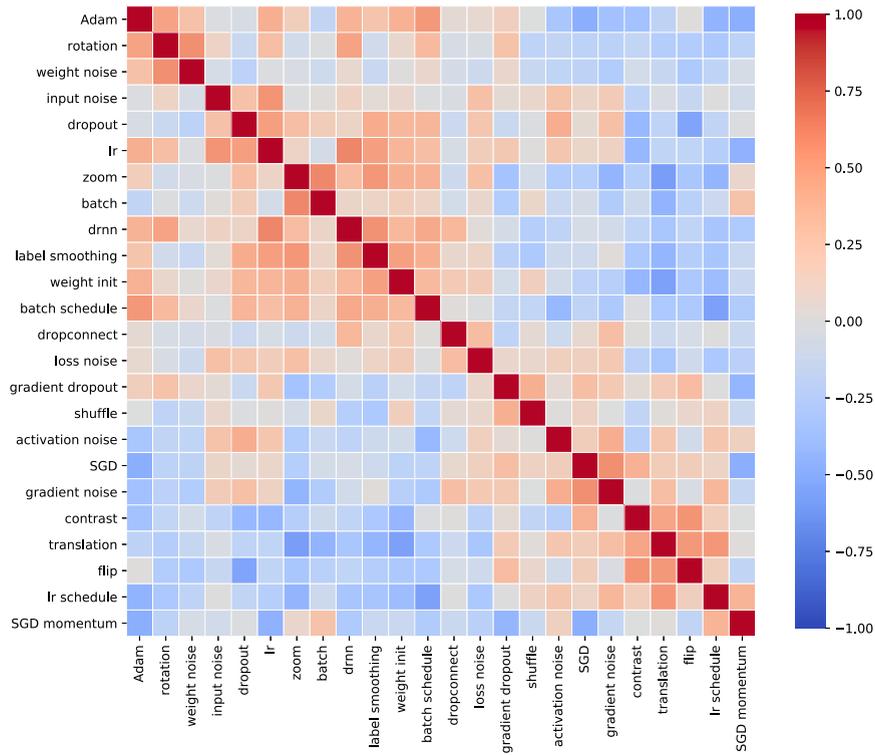

**Fig. 6.** The correlation heatmap capturing which randomness techniques were selected together among the 120 top-performing configurations.

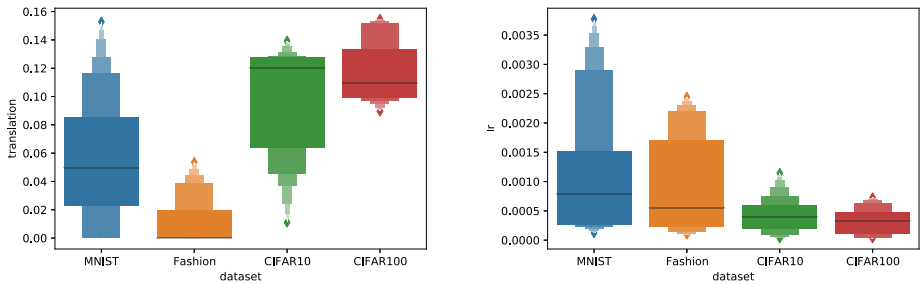

**Fig. 7.** The relation between the dataset and random translation (on the left) and learning rate (right).

### 5.4. Which optimizer to choose?

The associations conditioned by the choice of optimizer in Fig. 8 reveal that SGD works best with a large initial learning rate and small batch size. Both decisions contribute to a higher noise scale based on Equation (2). The strong correlations with gradient noise in the heatmap in Fig. 6 indicate that SGD further increasing the noise in the gradient estimate. During the ablation process, we observed that SGD does not perform well with large batch sizes, as using SGD instead of one of the other two optimizers led to training instabilities in settings with batch sizes larger than 512.

SGD with Momentum demonstrates entirely different preferences in terms of randomness compared to regular SGD, favoring a small starting learning rate and a large batch size. This combination reduces the noise scale according to Equation (2). This tendency could be explained by the trinity interplay [8] between the learning rate, the batch size, and the momentum according to Equation (2). The use of 0.9 momentum results in a 10 times larger noise scale. This effect prompts SGD with Momentum to prefer larger batch sizes, effectively compensating for the increase in noise. This preference explains why SGD with Momentum has been the preferred choice for training studies with larger batch sizes [8].

Adam shows similarity to SGD with the preference for large learning rates and small batch sizes. This combination maximizes the noise scale, resulting in large, noisy parameter updates. However, Adam differs in its tendency to decay the learning rate and increase the batch size more rapidly compared to the other two optimizers, as depicted in Fig. 8. This behavior leads to smaller, more accurate updates as the training progresses. This observation is interesting as it contradicts a common intuition in the deep learning community, which suggests that decaying the learning rate is not as important for adaptive optimizers due to their ability to use different step sizes for different parameters [47]. Our finding suggests the opposite, with Adam showing a preference for a more aggressive learning rate and batch schedules.

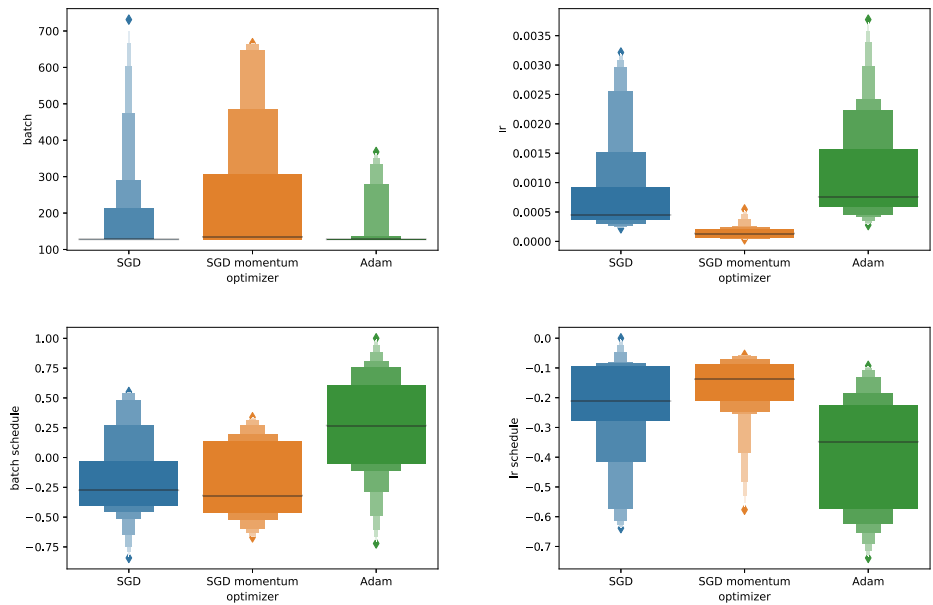

**Fig. 8.** The relation between the optimizer and batch size (top left), learning rate (top right), batch size schedule (bottom left), and learning rate schedule (bottom right).

In summary, SGD works best by maximizing the noise in the gradient estimates through a large learning rate and a small batch size. On the other hand, SGD with Momentum aims to balance the increased noise level induced by momentum by working with small updates derived from accurate gradient estimates. Finally, Adam starts with a large noise scale, similar to SGD, but quickly reduces it by increasing the batch size and decaying the learning rate.

We recommend accordingly confining the use of Adam to easier datasets with a limited computational budget for tuning the learning rate. In our experiments, Adam was only successful in 4 settings with the easier datasets MNIST and Fashion MNIST. The advantage of Adam lies in its capability to handle a broad spectrum of learning rates and weight initialization values, thereby reducing the necessity to fine-tune these hyperparameters. The limited success of Adam in our experiments aligns with a growing body of research that observed a degradation in generalization performance due to the use of adaptive optimizers [48].

SGD with Momentum demonstrated twice the success rate of Adam, being chosen in 8 instances. Furthermore, it exhibited greater consistency by being selected at least once for each dataset. The capability of SGD with Momentum to handle larger batches could potentially reduce training time by requiring fewer parameter updates. The use of large batches is, of course, dependent on the available memory of the employed hardware.

Finally, our results confirm SGD as the *de facto* optimizer of choice [47]. It was selected 7 times out of 10 for the harder datasets, i.e., CIFAR10 and CIFAR100. The very best settings for both hard datasets included SGD. This finding is consistent with the use of SGD for training many popular models like ResNet, Wide Residual Networks (WRNs), and DenseNet. However, in scenarios where training time is a concern, we recommend using SGD with Momentum, as the preference of SGD for small batch sizes is likely to result in significantly longer training times. Another alternative for mitigating the burden of a long training time is to use larger batch sizes with SGD while simultaneously increasing the noise in optimization using gradient noise. Similar proposals are suggested in [49] through adding covariance noise to the gradients. Another scenario where the use of SGD with Momentum could be beneficial is in the transfer learning setting. Given SGD with Momentum's preference to learn with a small learning rate, this could be useful for fine-tuning a pre-trained model by incrementally adapting the pre-learned features to the data. Besides, model convergence issues related to gradient explosion are reported with few-shot learning using large learning rates.

### 5.5. Why are loss noise and gradient dropout useful?

The results in Fig. 5 show that our two proposals consistently improved performance in a variety of scenarios. This raises an interesting question: why are these randomness techniques useful? We examine the training and validation curves qualitatively for different scenarios from our ablation experiments to answer this question. We notice that loss noise demonstrates a regularizing effect that prevents the network from memorizing the data and learning overly simple patterns early in the training.

According to Fig. 9, in both scenarios (MNIST and Fashion MNIST), the training curves in the early stages of the training process remain lower when using loss noise, compared to the training curves without loss noise. Moreover, while the validation curves with loss noise start similarly (slightly worse) to curves without loss noise, they reach consistently higher levels as the training progresses. In the end, they end up generalizing better. We conclude that the slow early progress in training is an indicator that the network with loss noise starts by learning more complex patterns, which consequently leads to higher final performance. We observe similar training and validation patterns in Fig. 10, by comparing configurations with gradient dropout and without it.

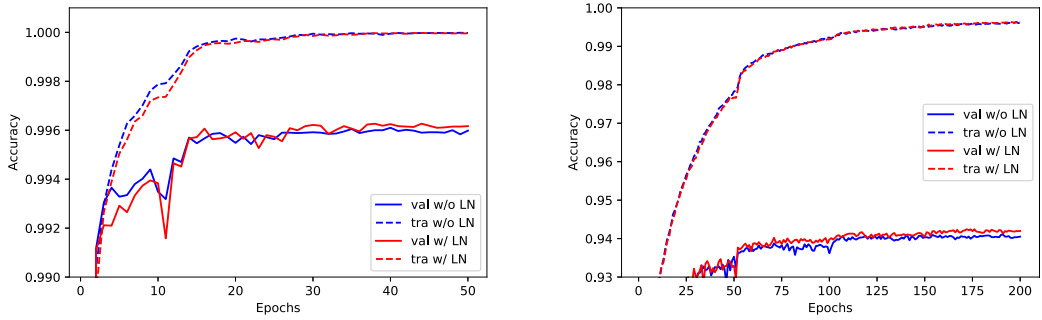

**Fig. 9.** The training and validation curves with and without loss noise for the MNIST (on the left) and Fashion MNIST (right) datasets.

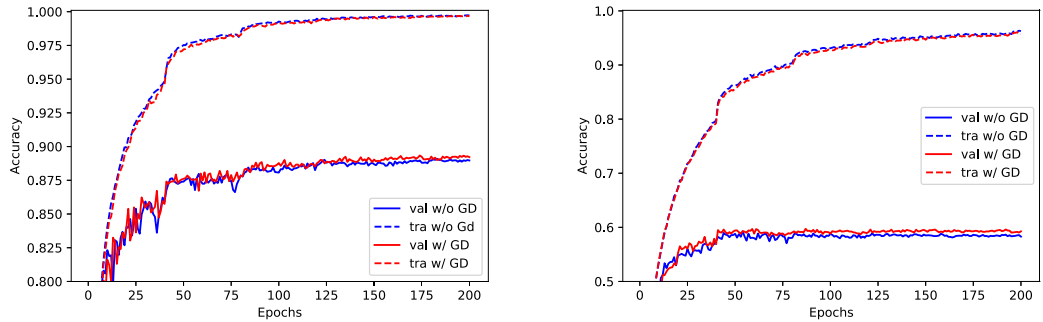

**Fig. 10.** The training and validation curves with and without gradient dropout for the CIFAR10 (on the left) and CIFAR100 (right) datasets.

## 6. Conclusions

This paper studies the compatibility and effectiveness of various randomness techniques in training DNNs. We categorize existing randomness techniques into types: data randomness, model randomness, optimizer randomness, and learner randomness, and discuss them based on their purposes: regularization, data size, convergence, and training time.

We examine which combinations of randomness techniques contribute most to the training performance of DNNs. Overall, the data randomization had the highest impact on performance; however, the significance of model and optimizer randomness is also evident with the consistent test error reductions of weight initialization, initial learning rate, and learning rate decay. The performance is greatly influenced by interactions among different randomization techniques, with the optimizer playing a key role in determining which other techniques are beneficial.

The analysis of the top-performing settings suggests that SGD with Momentum should be the optimizer of choice for large-batch training and few-shot learning scenarios. On the other hand, Vanilla SGD offered the highest performance for the harder datasets making it a natural choice for scenarios where performance is the main objective. However, hyperparameter selection is more crucial for Vanilla SGD, given its sensitivity to initialization, learning rate, and batch size. We would advise practitioners aiming to achieve the highest performance to avoid the use of Adam. Our study indicates that Adam's generalization performance falls significantly short compared to the SGD family of optimizers, especially for more challenging datasets. The upside of Adam is its capability to accommodate a broader range of learning rates and weight initialization values, which could be invaluable when computational resources for tuning these parameters are limited.

We propose two novel techniques, loss noise, that injects noise to loss function calculations, and gradient dropout that masks some of the gradients during optimization. Our proposals outperformed multiple well-established existing techniques (including the increasing of the batch size, gradient noise, and weight noise). They consistently improved the performance across various scenarios, matching the effectiveness of popular methods like dropout and label smoothing (see Fig. 5 for details). It must be noted, additionally, that our experimental setup is inherently more challenging compared to a conventional setup, in which the effectiveness of the intervention is measured while controlling all other interventions. In our setup, our methods are competing against 20 other randomness techniques.

Our result showed a superior performance of DropConnect compared to the most popular variant, Dropout, with fully connected layers. However, Dropout performed better with convolutional layers. We recommend accordingly, to use Dropout with the feature extraction layers, and DropConnect for the fully connected classification layers. This finding suggests that DropConnect should also be the preferred method for few-shot and transfer learning scenarios where a pre-trained feature extractor is used.

## 7. Future work and limitations

Our empirical findings are conditioned by the choice of standard architectures and low-resolution "easy" datasets, characterized by specific properties combining high dimensionality and class compactness. Our conclusions could be strengthened further by running the procedure with more datasets and architectures. The main difficulty lies in the vast search space, and the need to evaluate thousands of randomness configurations during optimization. It would be computationally infeasible to run the search for a high-resolution image dataset, e.g., ImagNet. But it could still be possible to run the search for a fine-tuning scenario, e.g., a ResNet50 network pre-trained with ImageNet, where we run the optimization for the downstream task, highlighting recommendations for transfer and few-shot learning settings. This scenario could be useful for deep learning practitioners, given that pre-trained models are almost always used in practical applications of DNNs.

In this paper, we have focused on image datasets, but this kind of study is relevant to other types of data like natural language, time series, and tabular data. It would be interesting to verify if our findings extend to other types of data, as the relative importance of randomness techniques could differ from image to time series data, for instance. One such example is data shuffling that is known to improve the performance of LSTM with time series data, in contrast, in our results we observed limited success with shuffling images. Also, the relative importance of data augmentation could very well differ for other data types. It is not unthinkable to find out that model and optimization techniques are more effective than data techniques for tabular data where the notion of augmentation is very data-specific.

The real-world applications of DNNs rarely focus solely on a single objective as a measure of success. Other metrics, like training and inference time, could be as important for the application as the accuracy. This is particularly important given that many of the randomization techniques were proposed to improve metrics beyond accuracy (e.g., Randomized Neural Networks were proposed to reduce training time by simplifying optimization). In this study, we solely evaluated the randomness approaches from a single objective (accuracy) point of view. It would be interesting to extend this work to multi-objective settings with two (or more) conflicting goals like performance and robustness. The study of another objective, such as convergence, could be equally valuable. For instance, investigating combinations that allow the training of deeper networks. Some techniques like [12] have demonstrated success (in isolation) in allowing training of very deep networks, and an extension of our work could discover more complex interactions between settings.

The high computation time resulting from the need to fully train a DNN to evaluate a candidate setting can be eased using approximate fitness evaluations. We are working on an extension that considers guiding the search through approximate models [50]. This approach would enable us to efficiently explore the search space more extensively.

Another possibility to extend the work is to study the randomness techniques on a layer-granularity. Many techniques (e.g., dropout and weight initialization) could potentially perform differently based on the layer type (e.g., dense or convolutional), or their position in the network (e.g., input facing or in the middle layer). It is understandable that such analysis requires solving harder optimization problems with more dimensions, but our PSO approach showed promise in identifying interesting regions of the search space leading to significant performance gains.

Finally, an interesting question regarding noise is whether and how it should be decayed or changed during training. A number of methods included in our study recommend a scheduler that mostly decays the scale of the noise as training progresses. Recently, research has been recommending sharp drop schedules [8], such as step-function drops, over previously popular exponential decay. It would be interesting to incorporate the choice of schedule within an extension of our ablation approach to identify successful scheduler(s) of noise.

**CRediT authorship contribution statement**

**Mohammed Ghaith Altarabichi:** Writing – review & editing, Writing – original draft, Visualization, Validation, Resources, Project administration, Methodology, Formal analysis, Data curation, Conceptualization. **Sławomir Nowaczyk:** Writing – review & editing, Writing – original draft, Validation, Supervision, Project administration, Methodology, Funding acquisition, Formal analysis. **Sepideh Pashami:** Writing – review & editing, Supervision, Project administration, Methodology, Funding acquisition. **Peyman Sheikholharam Mashhadi:** Writing – review & editing, Supervision. **Julia Handl:** Writing – review & editing, Writing – original draft, Visualization, Validation, Methodology.

**Declaration of competing interest**

The authors declare that they have no known competing financial interests or personal relationships that could have appeared to influence the work reported in this paper.

**Data availability**

We are sharing a GitHub repo with code and data.